\documentclass[runningheads]{llncs}
\usepackage[T1]{fontenc}
\usepackage{tabularray}

\usepackage{graphicx}
\usepackage{booktabs}
\usepackage{color}
\usepackage{amsmath}
\usepackage{amsfonts}
\usepackage{subfig}
\usepackage{bmpsize}
\usepackage{multirow}
\usepackage{float}

\usepackage{color}

\usepackage{pifont}
\usepackage[table]{xcolor}
\newcommand{\cmark}{\ding{51}}%
\newcommand{\xmark}{\ding{55}}%

\begin{document}
\title{Anticipating Future Object Compositions \\without Forgetting}

%
%
\author{Youssef Zahran\inst{1,2}\orcidID{0009-0008-5386-5530} \and
Gertjan Burghouts\inst{2}\orcidID{0000-0001-6265-7276} \and
Yke B. Eisma \inst{1}\orcidID{0000-0003-3437-2761}}
\authorrunning{Y.Zahran et al.}
%
\institute{Department of Cognitive Robotics, Delft University of Technology \and
Intelligent Imaging Department, TNO 
}
\maketitle              
\begin{abstract}
Despite the significant advancements in computer vision models, their ability to generalize to novel object-attribute compositions remains limited. Existing methods for Compositional Zero-Shot Learning (CZSL) mainly focus on image classification. This paper aims to enhance CZSL in object detection without forgetting prior learned knowledge. We use Grounding DINO and incorporate Compositional Soft Prompting (CSP) into it and extend it with Compositional Anticipation. We achieve a $70.5 \%$ improvement over CSP on the harmonic mean (HM) between seen and unseen compositions on the CLEVR dataset. Furthermore, we introduce Contrastive Prompt Tuning to incrementally address model confusion between similar compositions. We demonstrate the effectiveness of this method and achieve an increase of $14.5 \%$ in HM across the pretrain, increment, and unseen sets. Collectively, these methods provide a framework for learning various compositions with limited data, as well as improving the performance of underperforming compositions when additional data becomes available.
\end{abstract}

\section{Introduction}\label{section:Introduction}
Although humans have never seen a blue apple, they can easily picture it. This is due to the inherent human ability to generalize to novel concepts by combining the known entity "apple" with the color "blue". However, do computer vision models possess this capability? This question has motivated the development of Compositional Zero-Shot Learning (CZSL) \cite{RL-CZSL,HPL,CaNet,CAILA}. In CZSL, the goal is to recognize unseen object-attribute combinations, referred to as compositions, based on the compositions seen during training.
For this, models should understand the attributes and objects that compose these compositions to generalize to all possible compositions. 

Vision Language Models (VLMs), pretrained on large-scale image-text pairs, are promising for CZSL due to their ability to understand the relationship between the visual content and the textual description \cite{CLIP,CaNet,CAILA}. For object detection, VLMs such as Grounding DINO \cite{GroundingDino} and GLIP \cite{GLIP} learn to associate regions of text with regions of images by pulling the embeddings of paired image regions and text descriptions close while pushing others away \cite{vlmsurvey}. These models perform cross-modality fusion throughout the whole architecture, which makes the textual features image-aware and the visual features text-aware. Liu et al. \cite{GroundingDino} argue that VLMs benefit from frequent cross-modality fusion, making Grounding DINO superior to GLIP. Therefore, throughout this paper, we will solely focus on Grounding DINO. 
\raggedbottom

Unfortunately, these models tend to be biased towards object categories rather than attributes, which makes it suffer from feature misalignment when used directly for attribute recognition \cite{Chen_2023_CVPR}. Fine-tuning VLMs can solve this but often leads to catastrophic forgetting of prior knowledge \cite{zhou2023learningVLM}, thereby compromising their generalization ability. To address this, we explore how to fine-tune VLMs to perform well in CZSL without forgetting any prior knowledge. Nayak et al. \cite{CSP} introduced Compositional Soft Prompting (CSP), which combats catastrophic forgetting by adding auxiliary tokens for all words in a given dataset and training only these tokens. This approach preserves the model’s original embeddings, allowing it to retain and revert to its initial knowledge when necessary, unlike full fine-tuning, which alters the model’s parameters. CSP improves model performance in CZSL for image classification. We incorporate CSP in Grounding DINO, to leverage it for object detection.

We consider CSP as a baseline and improve it for CZSL by introducing Compositional Anticipation (CA), which recognizes that additional compositions may exist beyond those present during training. In this context, the term "anticipation" does not refer to actively predicting new compositions. Instead, it involves enhancing the model's ability to handle potential new compositions by adjusting how it processes partially correct predictions through Compositional Smoothing and by guiding the model to disentangle attributes from objects via Compositional Independence. Compositional Smoothing prepares the model for novel compositions by assigning soft labels when predictions are partially correct, e.g., the object is correct but the attribute is different. This approach deviates from conventional Label Smoothing \cite{labelsmoothing}, which assigns soft labels to all classes. Compositional Independence disentangles objects from attributes through Separation and Decorrelation. Separation introduces a separation loss to maximize the distinction between object and attribute classes by applying intra-class separation within objects and attributes and an inter-class separation between objects and attributes. Decorrelation minimizes the correlation between objects and attributes to reduce dependency between the two.

For incremental learning on newly added compositions, we use prior knowledge to address specific mistakes related to confusion between similar compositions. Inspired by recent developments in prompt tuning \cite{contextawareprompting,CoCOOP,COOP}, we introduce a novel method called Contrastive Prompt Tuning, specifically tailored for object attributes. Contrastive Prompt Tuning addresses cases where the model confuses similar compositions, such as mistaking a blue apple for a red apple, by adding a trainable prompt in front of the confused class: "is not red apple but is blue apple". This approach utilizes our prior knowledge to harness the ability of a VLM to exploit language.

In summary, our main contributions are:
\begin{enumerate}
    \item We incorporate CSP \cite{CSP} into Grounding DINO \cite{GroundingDino} and extend it with Compositional Anticipation. Compositional Anticipation consists of:
    \begin{itemize}
        \item Compositional Smoothing, which assigns soft labels when predictions are partially correct.
        \item Compositional Independence, which disentangles  objects from attributes.
    \end{itemize}
    \item We develop Contrastive Prompt Tuning, a method that adds a learnable prompt for compositions that are confused with each other during training. This technique harnesses the power of language and our understanding of the model to improve performance beyond simply training with additional data.
\end{enumerate}
\section{Related Work}\label{section:Related Work}
In this section, we review literature related to our work. We cover Compositional Zero-Shot Learning (CZSL), Prompt Tuning in Vision-Language Models (VLMs), and Class Incremental Learning (CIL). Our research focuses on improving the CZSL capabilities of Grounding DINO \cite{GroundingDino}, a VLM designed for object detection, by utilizing prompt tuning. Additionally, we address underperforming compositions in a class-incremental manner to further improve model performance.
\subsubsection{Compositional Zero-Shot Learning}
The main objective of CZSL is to recognize unseen compositions from the compositions encountered during training. In CZSL, individual objects and attributes are referred to as primitives. Misra et al. \cite{misra2017red} use a limited set of compositions to learn linear classifiers for each primitive. Then, they learn a transformation network that takes these classifiers as input and
composes them to produce a classifier for their combination. Since then, multiple works \cite{li2022siamese,Mancini_2021_CVPR,nagarajan2018attributes,saini2022disentangling} have been proposed to tackle the CZSL task. 

Recent works focus on adapting pretrained VLMs for CZSL by fine-tuning primitive tokens. While CSP \cite{CSP} only trains these tokens, others \cite{lu2023drpt,HPL} also introduce prompt disentangled tuning. This technique addresses entanglement, where optimizing one primitive's embedding affects another. Prompt disentangled tuning divides the process into three phases with different prompts: one for the entire composition, one for the attribute, and one for the object. This ensures attributes and objects learn their optimal parameters independently. 

While \cite{lu2023drpt,HPL} improve upon \cite{CSP} with an average performance increasement of $1.7 \%$ and $2.3 \%$, respectively, the gains are marginal relative to the increased complexity. Our work is closely related to \cite{lu2023drpt,CSP} as we adapt CSP for object detection and address entanglement through Compositional Independence.

\subsubsection{Prompt Tuning in VLMs}
Ever since CLIP \cite{CLIP} demonstrated that prompt templates such as “a photo of a [CLASS]” improve the results of VLMs compared to using only the classname, several other works \cite{contextawareprompting,CPT,CoCOOP,COOP} have been introduced to replace the hand-crafted prompt with learnable soft prompts. CoOP \cite{COOP} introduces soft prompts that are shared across all classes, resulting in prompts like \([v_1], [v_2], \ldots [v_M]\) for all images. CoCoOp \cite{CoCOOP} improves upon this by proposing soft prompts that are image-conditioned, generating prompts such as \([v_1(x)], [v_2(x)], \ldots [v_M(x)]\) for each image \(x\). Building upon these advancements, Rao et al. \cite{contextawareprompting} use contextual information from the image to prompt the language model.

Our work is closely related to these works but is unique in its focus on improving the performance of confused compositions using learnable prompts that are initialized based on our knowledge of the model's errors.

\subsubsection{Class Incremental Learning}
Class-Incremental Learning (CIL) refers to learning new classes while retaining previously learned classes \cite{zhou2023deep,zhou2023learningVLM}. In typical CIL scenarios, learning occurs through a sequence of training tasks, each of which introduce new classes without any overlap
of the classes from previous tasks. The main challenge is avoiding catastrophic forgetting, where learning new classes leads to a loss of knowledge from previous tasks. Our approach bears resemblance to Blurry CIL \cite{bang2021rainbow,bang2022online}, where former classes can be revisited during training. Similarly, we train incrementally with underperforming compositions while allowing former compositions to be revisited. 
\section{Method}\label{section:Method}
\subsection{Problem Definition}\
\subsubsection{Compositional Zero-Shot Learning}
We follow \cite{HPL,CaNet,CAILA} and formalize the CZSL task as follows. Let \( \mathcal{A} \) denote the set of attributes, and \( \mathcal{O} \) the set of objects, and \( \mathcal{C} = \mathcal{A} \times \mathcal{O} \) the set of all compositions. \( \mathcal{T} = \{(x_j, c_j)\}_{j=1}^{N} \) denotes the train set where \( x_j \in \mathcal{X} \) is a sample in the input (image) space \( \mathcal{X} \) and \( c_j \in \mathcal{C}_s \) is a composition in the subset \( \mathcal{C}_s \subseteq \mathcal{C} \). The seen set \( \mathcal{C}_s \subseteq \mathcal{C} \) consists of all compositions encountered during training, whereas the unseen set \( \mathcal{C}_u \subseteq \mathcal{C} \) consists of compositions not seen during training. Let \( \mathcal{C}_s\) and \( \mathcal{C}_u\) be two sets such that \( \mathcal{C}_s \cap \mathcal{C}_u = \emptyset \). While \(\mathcal{C}_s \) and \(\mathcal{C}_u \) are disjoint, the objects \(\mathcal{O}_u \) and attributes \(\mathcal{A}_u \) are defined such that \(\mathcal{O}_u \subseteq \mathcal{O}_s \) and  \(\mathcal{A}_u \subseteq \mathcal{A}_s \)

\subsubsection{Catastrophic Forgetting}
VLMs, such as Grounding DINO \cite{GroundingDino}, are known for their ability to generalize well across diverse tasks due to the extensive and varied data used during pre-training. However, fine-tuning these models on a new dataset often compromises their generalization capability, as the rich features learned during pre-training are replaced by features specific to the new dataset. This can lead to catastrophic forgetting, where the model's performance on previously learned tasks significantly deteriorates. In the context of CZSL with VLMs, catastrophic forgetting is particularly problematic. While the model may perform well on the specific compositions present in the new dataset it was fine-tuned on, it risks becoming overly specialized. 
This specialization may result in a model that loses its ability to generalize to other compositions, objects, or concepts, and instead becomes exceptionally good at predicting the compositions seen during training. Such a limitation is especially undesirable for open-set object detectors like Grounding DINO, which are meant to recognize a wide range of concepts.

\subsection{Incremental CZSL}
In practical settings, models often encounter new data or need to improve performance on underperforming compositions after the initial training phase. To address this, we introduce an increment set \( \mathcal{C}_i \subseteq \mathcal{C} \) to CZSL. Let \( \mathcal{C}_p \) be the set used for the initial fine-tuning of the model, with \( \mathcal{C}_p = \mathcal{C}_s \). After introducing \( \mathcal{C}_i \), the set of seen compositions becomes \( \mathcal{C}_s = \mathcal{C}_p \cup \mathcal{C}_i \). The increment set \( \mathcal{C}_i \) consists of compositions introduced after the initial fine-tuning to improve performance on underperforming compositions. Improving these underperforming compositions with additional compositions is challenging because \( \mathcal{C}_p \) is designed to cover \( \mathcal{A} \) and \( \mathcal{O} \) with the minimum number of compositions. Extending \( \mathcal{C}_s \) with \( \mathcal{C}_i \) makes the attributes and objects in \( \mathcal{C}_i \) overrepresented in \( \mathcal{C}_s \), which can bias the model towards these attributes and objects. In this paper, we focus solely on improving performance on compositions \( c_j \in \mathcal{C} \) without extending the attribute set \( \mathcal{A} \) or the object set \( \mathcal{O} \).
\begin{figure}[h!]
    \centering
\includegraphics[width=1.0\textwidth]{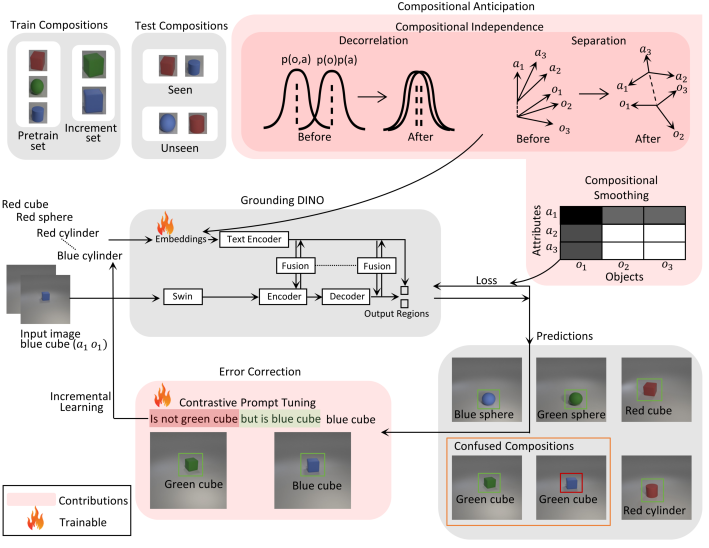}
\caption{Our method anticipates unseen, future object-attribute compositions through Compositional Independence and Compositional Smoothing. Forgetting is mitigated by creating auxiliary tokens for the language embeddings and refining only these tokens. Errors in compositions are incrementally corrected using Contrastive Prompt Tuning, which contrasts confused compositions.}
    \label{fig:method}
\end{figure}
\subsection{Compositional Soft Prompting}
To prevent catastrophic forgetting in Grounding DINO \cite{GroundingDino}, we follow CSP \cite{CSP} and modify it for object detection. Objects and attributes that form compositions are treated as learnable tokens within the VLMs vocabulary. 
Each attribute \(a_j \in \mathcal{A}\) and each object \(o_j \in \mathcal{O}\) is represented as an auxiliary token \(t_{a_j} \) and \(t_{o_j} \) respectively, where \(t_{a_j} , t_{o_j}  \in \mathbb{R}^d\), with \(d\) being the dimension of the vocabulary embedding. During training, only these auxiliary tokens are tuned, resulting in \((|\mathcal{A}| + |\mathcal{O}|) \times d\) learnable parameters. 

To illustrate, CSP creates auxiliary tokens for each attribute and object such as \(t_{blue}\) for the attribute "blue" and \(t_{apple}\) for the object "apple". These tokens are adjusted during training while the rest of the weights, such as those of the encoder and decoder in Grounding DINO, remain unchanged. By doing this, CSP prevents catastrophic forgetting and preserves the pretrained weights of the model.

\subsection{Compositional Smoothing}
To combat bias during training for \(\mathcal{C}_s\), where the model becomes overly confident with the seen classes, we assign soft labels rather than hard labels (0 and 1) in the classification loss. This is referred to as Label Smoothing  \cite{labelsmoothing}, and it prevents the model from becoming overly confident in its predictions, thereby improving its generalization capability. 
Conventional Label Smoothing adjusts the target labels by distributing a small portion of the probability mass to all other labels. For a given true label \(y\) in a classification problem with \(k\) classes, the smoothed label \(y_{\text{smooth}}\) is defined as:

\begin{equation}
y_{\text{smooth}} = (1 - \epsilon) y + \frac{\epsilon}{k},
\end{equation}
where \(\epsilon\) is the smoothing parameter, and the term \(\frac{\epsilon}{k}\) distributes the smoothing equally among all classes.

We deviate from conventional Label Smoothing \cite{labelsmoothing} and assign soft labels based on the correctness of the object, attribute, or the entire composition. We refer to this as Compositional Smoothing. Let \( p_{\mathcal{O}} \), \( p_{\mathcal{A}} \), and \( p_{\mathcal{C}} \) represent the probabilities for object, attribute, and overall composition predictions, respectively. For a given true composition \( c_t \) composed of object \( o_t \) and attribute \( a_t \), and predicted composition \( c_p \) composed of object \( o_p \) and attribute \( a_p \), the smoothed label \( y_{\text{o,a}} \) is defined as:

\begin{equation}
y_{\text{o,a}} = 
\begin{cases} 
p_{\mathcal{O}} & \text{if } o_p = o_t \land a_p \neq a_t, \\
p_{\mathcal{A}} & \text{if } a_p = a_t \land o_p \neq o_t, \\
p_{\mathcal{C}} & \text{if } a_p = a_t \land o_p = o_t, \\
0 & \text{otherwise}.
\end{cases}
\end{equation}
Compositional Smoothing ensures that there is a difference between having partial correctness and no correctness in the prediction, guiding the model to learn what the compositions are composed of rather than learning the compositions themselves. This, in turn, should lead to better performance on \(\mathcal{C}_u\).

Figure \ref{fig:method} (top right) illustrates Compositional Smoothing. For a given ground truth label $(a_1,o_1)$, predictions where both the attribute and object are correct are shown in black, and the smoothed label becomes $p_{\mathcal{C}}$. Partial correctness is depicted in gray, with the smoothed label being either $p_{\mathcal{O}}$ or $p_{\mathcal{A}}$. When both the object and attribute are completely wrong, no smoothing is applied.
\subsection{Compositional Independence}
In CZSL, it is important to disentangle objects from attributes and have clear distinctions withing each category. For example, a cube and a cylinder should be easily distinguishable to prevent confusion. Additionally, colors should be distinguished from specific objects, such as cubes, to ensure their independence. This prevents similar attributes or objects to be confused with each other and helps the model treat attributes and objects as distinct concepts. 

We achieve this independence through two components: Separation and Decorrelation. Separation enforces orthogonality within the embeddings of objects and attributes and maximizes the distance between their mean embeddings. Decorrelation minimizes the correlation between the embeddings of objects and attributes. This is achieved using the Hilbert-Schmidt Independence Criterion (HSIC) \cite{HSIC}, a kernel statistical test commonly used to measure independence between two random variables, which proved to be effective for CZSL image classification \cite{ruis2021independent} and is leveraged here for object detection.

\subsubsection{Separation}
To help the model differentiate between similar attributes or objects, we introduce an orthogonality loss. We achieve orthogonality within the groups of attributes and objects by minimizing the average absolute similarity between the normalized embeddings within each group: 
\begin{equation}
\mathcal{L}_{\text{orth}}(\mathbf{E}) = \frac{1}{|\mathbf{E}|^2 - |\mathbf{E}|} \sum_{i=1}^{|\mathbf{E}|} \sum_{\substack{j=1 \\ j \neq i}}^{|\mathbf{E}|} \left| \mathbf{e}_i \cdot \mathbf{e}_j \right|
\end{equation}
where \(\mathbf{E}\) is the set of normalized embeddings, and \(\mathbf{e}_i\) and \(\mathbf{e}_j\) are embeddings within this set. The summation \(\sum_{\substack{j=1 \\ j \neq i}}\) ignores self-similarity, and \(\frac{1}{|\mathbf{E}|^2 - |\mathbf{E}|}\) ensures that self-similar terms are excluded during normalization. This orthogonality loss is applied to both the attributes and objects:
\begin{equation}
\mathcal{L}_{\mathcal{A}} = \mathcal{L}_{\text{orth}}(\mathbf{E}_{\mathcal{A}})
\end{equation}
\begin{equation}
\mathcal{L}_{\mathcal{O}} = \mathcal{L}_{\text{orth}}(\mathbf{E}_{\mathcal{O}})
\end{equation}
where \(\mathbf{E}_{\mathcal{A}}\) and \(\mathbf{E}_{\mathcal{O}}\) represent the sets of normalized embeddings for the attributes and objects, respectively.

Additionally, to enforce a clear distinction between attributes and objects, we ensure that the mean embeddings of attributes and objects are significantly separated:
\begin{equation}
\mathcal{L}_{\text{distance}} = -\log(\|\mu_{\mathcal{A}} - \mu_{\mathcal{O}}\|_2)
\end{equation}
where \(\mu_{\mathcal{A}} = \frac{1}{|\mathcal{A}|} \sum_{j=1}^{|\mathcal{A}|} \mathbf{e}_{\mathcal{A}_j}\) and \(\mu_{\mathcal{O}} = \frac{1}{|\mathcal{O}|} \sum_{j=1}^{|\mathcal{O}|} \mathbf{e}_{\mathcal{O}_j}\) represent the mean embeddings of attributes and objects, respectively. The distance is computed using the \(L_2\) norm between the mean embeddings of the two groups.

The total Separation loss is a weighted combination of the orthogonality and mean separation components:
\begin{equation}
\mathcal{L}_{\text{separation}} = \lambda_1 \mathcal{L}_{\text{distance}} + \lambda_2 \mathcal{L}_{\mathcal{A}} + \lambda_3 \mathcal{L}_{\mathcal{O}}
\end{equation}
where \(\lambda_1\),\(\lambda_2\) and \(\lambda_3\) are hyperparameters
controlling the contribution of \( \mathcal{L}_{\text{distance}}\),
\(\mathcal{L}_{\mathcal{A}}\) and \(\mathcal{L}_{\mathcal{O}}\) to the final loss, respectively.

\subsubsection{Decorrelation}
To further ensure the independence between object and attribute embeddings, we introduce Decorrelation by using HSIC \cite{HSIC}. For an object \(o_j\) with attribute \(a_j\), we formulate the HSIC loss as follows:
\begin{equation}
\mathcal{L}_{\text{hsic}} = \lambda_h \text{HSIC}(o_j, a_j)
\end{equation}
Here, \(\lambda_h\) is a hyperparameter that controls the contribution of the HSIC term to the total loss.
\subsection{Compositional Anticipation}
Our method, which we refer to as Compositional Anticipation (CA), consists of both Compositional Smoothing and Compositional Independence. While CA does not actively predict unseen compositions, it prepares the model by refining how it handles potential new compositions and disentangles attributes from objects. Figure \ref{fig:method} shows how we implement CA in Grounding DINO \cite{GroundingDino}. 

\subsection{Contrastive Prompt Tuning}
To improve the performance of some underperforming composition after training with \( \mathcal{C}_p \), we extend \( \mathcal{C}_s \) with an additional set \( \mathcal{C}_i \) to improve performance. Our approach begins with analyzing the predictions to identify compositions that are frequently confused with each other. For instance, if \( c_j \) and \( c_k \) are often mixed-up, both compositions are included in \( \mathcal{C}_i \), and a trainable prompt is added in front of the underperforming class(es). For example, if \( c_j \) performs poorly, we add the following prompt in front of the class: \textit{"is not \( c_k\) but is \( c_j\)"}. This prompt contains both a negative and an affirmative component.

We refer to this method as Contrastive Prompt Tuning and it does not modify any of the tokens present in the sets \(\mathcal{A}\) and \(\mathcal{O}\). Instead, it focuses solely on the learnable prompt, which leads to fewer changes in the performance of other compositions and mitigates catastrophic forgetting. By doing this, we exploit the ability of a VLM to understand language and use a semantically meaningful initial prompt to learn to distinguish between similar compositions. This step is depicted as Incremental Learning in Figure \ref{fig:method}.

\section{Experiments}\label{section:Experiments}
\subsection{Evaluation}
\subsubsection{Dataset}
We evaluate our approach using a synthetic dataset generated following the CLEVR framework \cite{clevrpaper}. This dataset consists of three  types of objects: cube, cylinder, and sphere. Each object is associated with six attributes: blue, red, green, purple, brown, and yellow.

The dataset intentionally excludes non-visual attributes (e.g., heavy) and attributes that exhibit significant variation across different objects (e.g., wet in wet dog versus wet car). This yields a dataset that is reliable for assessing a model's performance in the CZSL task. Given that there are no ambiguous attributes present in this dataset, a poorly performing model would indicate that the model is bad in the CZSL task.
\subsubsection{Train-Test Split}
Throughout this section, all experiments for the CZSL task are trained using the set: $\{$red cube, blue cube, green sphere, purple sphere, brown cylinder, yellow cylinder$\}$ as \(\mathcal{C}_p\) with 10 shots per composition. This split ensures that \(\mathcal{C}_s\) covers the entire set of objects \(\mathcal{O}\) and attributes \(\mathcal{A}\). Testing is performed with the whole set of composition \(\mathcal{C}\) with 60 samples per composition.
\subsubsection{Evaluation Metric} 
We adopt the NMS mAP evaluation metric introduced by Yoa et al. \cite{nmsMAP}. In this work they argued that the traditional COCO mAP \cite{COCO} is deceiving for open vocabulary detection models, such as Grounding DINO \cite{GroundingDino}. Consider an image annotated with two ground-truth instances: a purple cylinder and a green
cylinder, assuming these are the only cylinder categories
in the model. These models tend to be able to detect and
locate the presence of all cylinders in the image, but they
struggle with the contextual description. They would predict
two overlapping bounding boxes for each object, mistakenly
assigning both ’green’ and ’purple cylinder’ labels to each
object. All four of these boxes would be predicted with a
high confidence score. Additionally, the highest scoring label
is not necessarily the correct one. Consequently, the AP for
each category would misleadingly be 0.50, despite the model failing to correctly comprehend the target objects. Yao et al.
\cite{nmsMAP} refer to this as the ’inflated AP problem’. 

To address this issue, Yao et al. \cite{nmsMAP} propose applying class-agnostic Non-Maximum Suppression (NMS) before calculating the mAP. This method suppresses redundant bounding boxes, ensuring that only the prediction with the highest confidence score is used in the calculation of the mAP. We adopt this NMS mAP metric to provide a more realistic measure of our model's performance.

\subsection{CSP base} 
We adapt CSP \cite{CSP} and modify it for Grounding DINO \cite{GroundingDino} and this integration serves as our baseline method. To assess its performance, we begin by training it with \(\mathcal{C}_p = \mathcal{C}\). This yields an NMS mAP of \(87.2 \pm 6.8\), demonstrating that good performance can be achieved by only training the embeddings of \(\mathcal{O} \) and \(\mathcal{A} \).

\begin{table}[b!]
\caption{Compositional Anticipation improves both object detection performance and generalization to unseen compositions. Compositional Smoothing contributes the most to these improvements, followed by Separation and Decorrelation.}

\label{tab:experiment1_differentsettingsResults}
\resizebox{\textwidth}{!}{
\begin{tabular}{c@{\hskip 0.5cm} c@{\hskip 0.5cm} c @{\hskip 0.5cm}c@{\hskip 0.5cm}c@{\hskip 0.5cm}c@{\hskip 0.5cm}c}
\toprule
\multicolumn{3}{c}{Compositional Anticipation (CA)} & & & & \\
\cmidrule(lr){1-3} 

Compositional Smoothing &  Separation &  Decorrelation  & Seen & Unseen  & HM \\ \midrule
\xmark & \xmark & \xmark & 81.4 $\pm$ 7.6 & \;\;4.5 $\pm$ 4.6 & \;\;8.0 $\pm$ 8.1 \\ 
\xmark & \xmark & \cmark & 81.3 $\pm$ 7.7 &  10.8 $\pm$ 6.7 & \; 18.2 $\pm$ 10.5 \\
\xmark & \cmark & \xmark & 82.5 $\pm$ 6.8 & 15.1 $\pm$ 4.2 & 25.4 $\pm$ 6.1 \\

\xmark & \cmark & \cmark & 84.4 $\pm$ 7.1 & 20.8 $\pm$ 4.7 & 33.1 $\pm$ 6.4 \\

\cmark & \xmark & \xmark & 86.2 $\pm$ 6.8 & 64.3 $\pm$ 5.9 & 73.5 $\pm$ 5.3 \\
\cmark & \xmark & \cmark & 92.4 $\pm$ 3.0 & 61.6 $\pm$ 5.7 & 73.8 $\pm$ 4.5 \\
\cmark & \cmark & \xmark & 86.0 $\pm$ 6.1 & 67.7 $\pm$ 4.7 &   75.7 $\pm$ 5.0 \\
\cmark & \cmark & \cmark & 88.7 $\pm$ 4.9 & 70.6 $\pm$ 7.4 & \textbf{78.5 $\pm$ 6.0} \\ 
\bottomrule
\end{tabular}
}
\end{table}
\begin{table}[b!]
\caption{Our model does not forget. It achieves good performance on the fine-tuned CLEVR \cite{clevrpaper} dataset while preserving performance on MS-COCO \cite{COCO}, whereas conventional fine-tuning of Grounding DINO \cite{GroundingDino} leads to forgetting on MS-COCO. }\label{tab:COCOcomparison}

\resizebox{\textwidth}{!}{
\begin{tabular}{l@{\hskip 0.5cm} c@{\hskip 0.5cm} c@{\hskip 0.5cm} c@{\hskip 0.5cm} c}
\toprule
& \multicolumn{2}{c}{CLEVR \cite{clevrpaper}} & \multicolumn{2}{c}{MS-COCO \cite{COCO}} \\
\cmidrule(lr){2-3} \cmidrule(lr){4-5}
\multicolumn{1}{c}{Model} & Before & After & Before & After \\ \midrule
Grounding DINO \cite{GroundingDino} & \multirow{2}{*}{23.4} & 91.0 \textcolor{blue}{$^{\uparrow 67.6}$} & \multirow{2}{*}{41.1} & 11.8 \textcolor{red}{$^{\downarrow 29.3}$} \\ 
+ CSP \cite{CSP} $+$  CA (ours)& & 76.6 \textcolor{blue}{$^{\uparrow 53.2}$} & & 41.1 \textcolor{blue}{$^{=  0.0}$} \\ 
\bottomrule
\end{tabular}
}
\end{table}

\subsection{CZSL Comparison}
We compare the CSP \cite{CSP} baseline with our proposed method, Compositional Anticipation (CA) which extends CSP with Compositional Independence and Compositional Smoothing. The results, averaged over 13 experimental runs, are shown in Table \ref{tab:experiment1_differentsettingsResults} and are denoted using the NMS mAP metric \cite{nmsMAP}. Our results show that our method substantially improves upon the CSP baseline, with the harmonic mean (HM) between seen and unseen compositions improving by $70.5 \%$. This improvement is predominately achieved on the unseen compositions, which improved by $66.1 \%$.

Additionally, we showcase that our method does not suffer from catastrophic forgetting by evaluating its generalization ability compared to the conventional fine-tuning of Grounding DINO \cite{GroundingDino}. We compare the results on the MS-COCO \cite{COCO} dataset before and after fine-tuning on the CLEVR \cite{clevrpaper} dataset for the CZSL task. Table \ref{tab:COCOcomparison} shows that conventional fine-tuning of Grounding DINO \cite{GroundingDino} achieves a $67.6 \%$ improvement on CLEVR, whereas our method achieves a $53.2 \%$ improvement. This suggests that conventional fine-tuning of Grounding DINO is superior in CZSL. However, conventional fine-tuning of Grounding DINO leads to a $29.3 \%$ performance drop on MS-COCO, whereas our method maintains stable performance with no drop at all. This demonstrates that conventional fine-tuning suffers from catastrophic forgetting, while our method does not.  

\subsection{Improving Incrementally}
In this experiment, we incrementally learn new classes using the model initially trained with CSP \cite{CSP} extended with Compositional Anticipation. We continue training the model using a dataset that includes both \(\mathcal{C}_p\) and \(\mathcal{C}_i\).

We explore two different fine-tuning methods: fine-tuning class-specific tokens and our proposed method, Contrastive Prompt Tuning. For fine-tuning class-specific tokens, we compare CSP \cite{CSP} with CSP extended with Compositional Anticipation. Additionally, we conduct this fine-tuning in two ways: (1) allowing the fine-tuning of all tokens in the sets \(\mathcal{O}\) and \(\mathcal{A}\), and (2) fine-tuning only objects and attributes present in \(\mathcal{C}_i\), specifically \(\mathcal{O}_i\) and \(\mathcal{A}_i\). For Contrastive Prompt Tuning, all tokens are frozen and only the prompt is fine-tuned. The prompt is initialized with semantically meaningful information, including both an affirmative and a negative component. For instance, if "green cylinder" is often confused with "green cube", the prompt is initialized as "is not green cube but is green cylinder". We also analyze the individual contributions of each component to the overall performance enhancements.

To evaluate performance, we compare the model's results before and after introducing \(\mathcal{C}_i\). Specifically, we determine performance across the sets \(\mathcal{C}_p\), \(\mathcal{C}_i\), and \(\mathcal{C}_u\). Initially, \(\mathcal{C}_u\) is defined as \(\mathcal{C} - \mathcal{C}_p\). After introducing \(\mathcal{C}_i\), \(\mathcal{C}_u\) becomes \(\mathcal{C} - \mathcal{C}_p - \mathcal{C}_i\). The results on these sets after introducing \(\mathcal{C}_i\) are shown in Table \ref{tab:incremental_results}, with the absolute changes compared to the initial values indicated with arrows. 

Our results show that our method, Contrastive Prompt Tuning, which fine-tunes a prompt initialized with prior knowledge to address specific mistakes related to confusion between similar compositions, is superior to the class-specific tuning strategy. With Contrastive Prompt Tuning, we achieve a $12.9 \%$ enhancement in the HM across the pretrain, increment, and unseen sets compared to the best class-specific tuning method. This improvement is predominately achieved across the increment set, which improved by $31.3 \%$ compared to the best class-specific tuning method. Furthermore, Contrastive Prompt Tuning benefits from both the affirmative and negative components of the prompt.
\begin{table}[t!]
\centering
\caption{Our Contrastive Prompt Tuning is effective for incremental learning. It improves performance across all classes including the unseen ones.}
\label{tab:incremental_results}
\resizebox{\textwidth}{!}{
\begin{tabular}{l c@{\hskip 0.25cm} c@{\hskip 0.25cm} c@{\hskip 0.25cm} c@{\hskip 0.25cm} c}
\toprule
\multicolumn{6}{c}{Class-Specific Tuning} \\ \midrule
\multicolumn{1}{c}{Method} & Tunable Tokens & Pretrained & Increment & Unseen & HM \\ \midrule
CSP & \(\mathcal{O} + \mathcal{A}\) & 88.1 $\pm$ 4.5 \textcolor{red}{$^{\downarrow 2.9}$} & \;72.9 $\pm$ 20.3 \textcolor{blue}{$^{\uparrow 13.6}$} & \;\; 0.0 $\pm$ 0.0 \textcolor{red}{$^{\downarrow 74.3}$} & \;\; 0.0 $\pm$ 0.0 \textcolor{red}{$^{\downarrow 71.7}$} \\ 
+ CA & \(\mathcal{O} + \mathcal{A}\) & \;80.9 $\pm$ 5.0 \textcolor{red}{$^{\downarrow 10.1}$} & 60.5 $\pm$ 18.2 \textcolor{blue}{$^{\uparrow 1.2}$} & 76.5 $\pm$ 8.4 \textcolor{blue}{$^{\uparrow 2.1}$} & 69.6 $\pm$ 7.0 \textcolor{red}{$^{\downarrow 2.1}$} \\ 
CSP & \(\mathcal{O}_i + \mathcal{A}_i\) & 89.0 $\pm$ 2.3 \textcolor{red}{$^{\downarrow 2.0}$} & 64.1 $\pm$ 20.9 \textcolor{blue}{$^{\uparrow 4.8}$} & 69.4 $\pm$ 8.3 \textcolor{red}{$^{\downarrow 4.9}$} & 70.6 $\pm$ 8.2 \textcolor{red}{$^{\downarrow 1.1}$} \\ 
+  CA & \(\mathcal{O}_i + \mathcal{A}_i\) & 89.2 $\pm$ 3.9 \textcolor{red}{$^{\downarrow 1.8}$} & 62.4 $\pm$ 19.6 \textcolor{blue}{$^{\uparrow 3.1}$} & 78.6 $\pm$ 4.9 \textcolor{blue}{$^{\uparrow 4.3}$} & 73.3 $\pm$ 9.1 \textcolor{blue}{$^{\uparrow 1.5}$} \\ \midrule
\multicolumn{6}{c}{Compositional Prompt Tuning (ours)} \\ \midrule
Affirmation & Prompt & 88.8 $\pm$ 5.5 \textcolor{red}{$^{\downarrow 2.2}$} & 82.8 $\pm$ 17.8 \textcolor{blue}{$^{\uparrow 23.5}$} & 74.8 $\pm$ 6.6 \textcolor{blue}{$^{\uparrow 0.5}$} & \;80.2 $\pm$ 7.9 \;\textcolor{blue}{$^{\uparrow 8.5}$} \\ 
Negation & Prompt & 91.2 $\pm$ 2.0 \textcolor{blue}{$^{\uparrow 0.2}$} & 81.9 $\pm$ 13.8 \textcolor{blue}{$^{\uparrow 22.6}$} & 76.6 $\pm$ 5.8 \textcolor{blue}{$^{\uparrow 2.3}$} & \;\;82.1 $\pm$ 6.0 \;\textcolor{blue}{$^{\uparrow 10.4}$} \\ 
Both & Prompt & 92.6 $\pm$ 1.5 \textcolor{blue}{$^{\uparrow 1.6}$} & 93.7 $\pm$ 2.1 \;\;\textcolor{blue}{$^{\uparrow 34.4}$} & 75.2 $\pm$ 5.0 \textcolor{blue}{$^{\uparrow 0.9}$} & \;\textbf{86.2 $\pm$ 2.1} \textcolor{blue}{$^{\uparrow 14.5}$} \\ 
\bottomrule
\end{tabular}
}
\end{table}

\begin{figure}[t!] 
    \centering
    \subfloat[\centering Before incremental learning
]{{\includegraphics[width=0.45\textwidth]{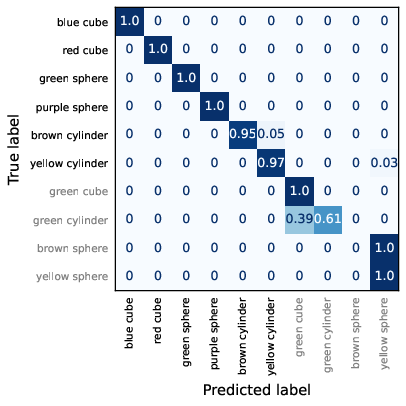}}}%
    \qquad 
    \subfloat[\centering After incremental learning]{{\includegraphics[width=0.45\textwidth]{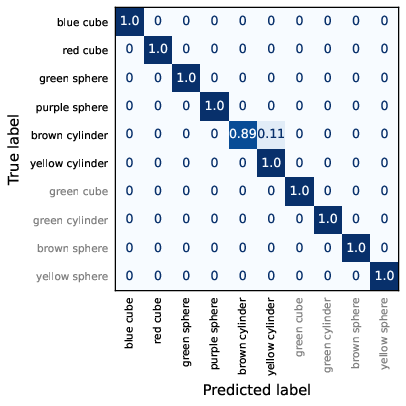} }}%
    
     \caption{Our Contrastive Prompt Tuning method is effective in incremental learning. It improves performance on the increment compositions (in gray) while preserving performance of the pretrained compositions (in black).}

    \label{fig:confusionmatrix}%
\end{figure} %

Figure \ref{fig:confusionmatrix} shows the effects of Compositional Prompt Tuning on the model's predictions. Figure \ref{fig:confusionmatrix}a shows that before incremental learning, 39\% of all instances of "green cylinder" are misclassified as "green cube", and all instances of "brown sphere" are misclassified as "yellow sphere". Figure \ref{fig:confusionmatrix}b demonstrates that after applying  Compositional Prompt Tuning, "green cube", "green cylinder", "yellow sphere", and "brown sphere" are classified correctly on all instances.

\section{Conclusion}\label{section:Conclusion}
In this paper,  we demonstrated that conventional fine-tuning of Grounding DINO achieves an NMS mAP of $91.0$ when fine-tuned on the CLEVR dataset for CZSL. However, this approach suffers from catastrophic forgetting, as confirmed by a $29.3 \%$ decrease in performance on the MS-COCO dataset post fine-tuning. To address this, we proposed incorporating CSP into Grounding DINO to mitigate forgetting by only fine-tuning auxiliary tokens. However, we observed that using CSP alone resulted in an NMS mAP of only $8.0$ for the HM between seen and unseen compositions. Therefore, we extended CSP with Compositional Anticipation, which improved the HM by $70.5 \%$. While our method improves upon the CSP baseline, it does not surpass conventional fine-tuning of Grounding DINO. 
Additionally, we introduced Contrastive Prompt Tuning to incrementally improve compositions that are confused with each other during training. With Contrastive Prompt Tuning, we improve performance on the HM across the pretrain, increment, and unseen sets by $12.9 \%$ compared to the best class-specific tuning method.

Given these findings, we recommend conventional fine-tuning of Grounding DINO for applications where performance on a specific dataset is prioritized, and our method for scenarios emphasizing overall performance across datasets. However, we acknowledge that our experiments are limited to the CLEVR dataset, and it remains unclear how the proposed methods will perform on real-world datasets beyond this toy dataset.

Considering that Grounding DINO excels in CZSL, likely due to the cross-modality fusion between image and text embeddings and our proposed methods involve strategically guiding the positioning of embeddings in the embedding space. Having demonstrated the benefits of our approach for CZSL, further investigation into the positioning of embeddings in the fused embedding space could potentially yield results approaching those achieved by conventional fine-tuning of Grounding DINO, but without encountering catastrophic forgetting.

\bibliographystyle{splncs04}
\bibliography{bibliograph.bib}

\end{document}